\documentclass[11pt,letterpaper]{article}
\usepackage{emnlp2016}
\usepackage{times}
\usepackage{latexsym}

\emnlpfinalcopy

\def\emnlppaperid{***}

\usepackage{latexsym}
\usepackage{graphicx}
\usepackage{multirow}
\usepackage{amsmath}
\usepackage{amssymb}
\usepackage{mathtools}
\usepackage{subcaption}

\title{How NOT To Evaluate Your Dialogue System: An Empirical Study of Unsupervised Evaluation Metrics for Dialogue Response Generation}
% ...

\author{\\\textbf{Chia-Wei Liu}$^{1}$\thanks{\hspace{1mm} Denotes equal contribution.}, \textbf{Ryan Lowe}$^{1*}$, \textbf{Iulian V. Serban}$^{2*}$,
\textbf{Michael Noseworthy}$^{1*}$,\\ \textbf{Laurent Charlin}$^1$, \textbf{Joelle Pineau}$^1$\\ 
	    $^1$ School of Computer Science, McGill University\\
	    {\tt \{chia-wei.liu,ryan.lowe,michael.noseworthy\}@mail.mcgill.ca} \\
	    {\tt \{lcharlin, jpineau\}@cs.mcgill.ca}\\
	    $^2$ DIRO, Universit{\'e} de Montr{\'e}al\\
	    {\tt iulian.vlad.serban@umontreal.ca}\\
 }
\date{}

\begin{document}

\maketitle

\begin{abstract}
We investigate evaluation metrics for dialogue response generation systems where supervised labels, such as task completion, are not available. Recent works in response generation have adopted metrics from machine translation to compare a model's generated response to a single target response. We show that these metrics correlate very weakly with human judgements in the non-technical Twitter domain, and not at all in the technical Ubuntu domain. We provide quantitative and qualitative results highlighting specific weaknesses in existing metrics, and provide recommendations for future development of better automatic evaluation metrics for dialogue systems.

%This work investigates a new class of metrics for evaluating the performance of dialogue systems, without requiring supervised (human-generated) task completion labels or user simulations. The metrics compute compositional embeddings over sentences, and measure dissimilarity between utterances using cosine distance. We use these metrics to evaluate the performance of recent retrieval and generative end-to-end dialogue models. 
\end{abstract}

\section{Introduction}

An important aspect of dialogue response generation systems, which are trained to produce a reasonable utterance given a conversational context, is how to evaluate the quality of the generated response. Typically, evaluation is done using human-generated supervised signals, such as a task completion test or a user satisfaction score \cite{walker1997paradise,moller2006memo,kamm1995user}, which are relevant when the dialogue is task-focused.
We call models optimized for such supervised objectives \textit{supervised dialogue models}, while those that do not are \textit{unsupervised dialogue models}.
%We call models that are trained to optimize for such supervised objectives \textit{supervised dialogue models}, while those that do not are \textit{unsupervised dialogue models}.

%Significant progress has been made in learning end-to-end systems directly from large amounts of text data for a variety of natural language tasks, such as question answering \cite{weston2015}, machine translation \cite{cho2014learning}, and dialogue response generation systems \cite{sordoni2015aneural}, in particular through the use of neural network models. 
%In the case of dialogue systems, an important challenge is to provide a reliable evaluation of the learned systems. Typically, evaluation is done using human-generated supervised signals, such as a task completion test or a user satisfaction score \cite{walker1997paradise,moller2006memo,kamm1995user}. We call models that are trained to optimize for such supervised objectives \textit{supervised dialogue models}, while those that are not \textit{unsupervised dialogue models}.

%While supervised models have historically been the favoured approach, supervised labels are difficult to collect on a large scale due to the cost of human labour. Further, for free-form types of dialogues (e.g., chatbots), the notion of task completion is ill-defined since it may differ from one human user to another. 
This paper focuses on unsupervised dialogue response generation models, such as chatbots. These models are receiving increased attention, particularly using end-to-end training with neural networks~\cite{serban2016hierarchical,sordoni2015aneural,vinyals2015neural}. This avoids the need to collect supervised labels on a large scale, which can be prohibitively expensive. 
%These models are typically trained (end-to-end) to predict the next utterance of a conversation, given several context utterances \cite{2015arXiv150704808S}. %This task is referred to as response generation. 
However, automatically evaluating the quality of these models remains an open question. Automatic evaluation metrics would help accelerate the deployment of unsupervised response generation systems.
%Having access to an automatic metric for evaluating unsupervised dialogue models would help accelerate the deployment of dialogue systems.

%, end-to-end dialogue systems have only been applied in the unsupervised case

Faced with similar challenges, other natural language tasks have successfully developed automatic evaluation metrics. For example, BLEU~\cite{papineni2002bleu} and METEOR~\cite{banerjee2005meteor} are now standard for evaluating  machine translation models, and ROUGE~\cite{lin2004rouge} is often used for automatic summarization. These metrics have recently been adopted by dialogue researchers \cite{ritter2011data,sordoni2015aneural,li2015diversity,galley2015deltableu,wen2015semantically,li2016persona}. However these metrics assume that valid responses have significant word overlap with the ground truth responses.
This is a strong assumption for dialogue systems, where there is significant diversity in the space of valid responses to a given context. This is illustrated in Table \ref{tab:toy}, where two reasonable responses are proposed to the context, but these responses do not share any words in common and do not have the same semantic meaning.

%Since the machine translation task %of correctly predicting a translation given an input sentence 
%appears similar to the dialogue response generation task, many dialogue researchers have adopted the same metrics for evaluating the performance of unsupervised dialogue  models~, even though these metrics have not been validated for dialogue-related tasks.  %unsupervised dialogue systems. 
% However, this is not necessarily valid; 
%Further, there is reason to believe these metrics may not transfer; 

%which are tasked with predicting a correct translated input utterance. 
%Unsupervised dialogue systems are similarly tasked with predicting the next utterance given a the previous utterances of a dialog, a \emph{context}. 
%In part because of this similarity, 

\begin{table}%[h]
\footnotesize
\centering
\begin{tabular}{l}
\hline
\textbf{Context of Conversation} \\ %\hline
Speaker A: Hey John, what do you want to do tonight?\\
Speaker B: Why don't we go see a movie?\\ \hline
\textbf{Ground-Truth Response}\\ %\hline
Nah, I hate that stuff, let's do something active. \\ %\hline
\textbf{Model Response}\\ %\hline
Oh sure! Heard the film about Turing is out!\\
%Reponse 3: It's warm out shall we try to the drive-in? \\ \hline
\hline
\end{tabular}
\caption{\label{tab:toy} Example showing the intrinsic diversity of valid responses in a dialogue. The (reasonable) model response would receive a BLEU score of 0.}%Even though the actual next response and the model generated response have the same sentiment and are both reasonable given the context, there is no word overlap between them.}
\end{table}

In this paper, we investigate the correlation between the scores from several automatic evaluation metrics and human judgements of dialogue response quality, for a variety of response generation models. We consider both statistical \emph{word-overlap similarity metrics} such as BLEU, METEOR, and ROUGE, and \emph{word embedding metrics} derived from word embedding models such as Word2Vec~\cite{mikolov2013distributed}. We find that all metrics show either weak or no correlation with human judgements, despite the fact that word overlap metrics have been used extensively in the literature for evaluating dialogue response models (see above, and Lasguido et al.~\shortcite{lasguido2014utilizing}). In particular, we show that these metrics have only a small positive correlation on the chitchat oriented Twitter dataset, and no correlation at all on the technical Ubuntu Dialogue Corpus.
For the word embedding metrics, we show that this is true even though all metrics are able to significantly distinguish between baseline and state-of-the-art models across multiple datasets.
We further highlight the shortcomings of these metrics using: a) a statistical analysis of our survey's results; b) a qualitative analysis of examples from our data; and c) an exploration of the sensitivity of the metrics. 

Our results indicate that a shift must be made in the research community away from these metrics, and highlight the need for a new metric that correlates more strongly with human judgement.

\section{Related Work}

We focus on metrics that are \textit{model-independent}, i.e.\@ where the model generating the response does not also evaluate its quality; thus, we do not consider word perplexity, although it has been used to evaluate unsupervised dialogue models~\cite{2015arXiv150704808S}. This is because it is not computed on a per-response basis, and cannot be computed for retrieval models. Further, we only consider metrics that can be used to evaluate proposed responses against ground-truth responses, so we do not consider retrieval-based metrics such as recall, which has been used to evaluate dialogue models~\cite{schatzmann2005quantitative,lowe2015ubuntu}. We also do not consider evaluation methods for supervised evaluation methods.\footnote{Evaluation methods in the supervised setting have been well studied, see~\cite{walker1997paradise,moller2006memo,jokinen2009spoken}.}

Several recent works on unsupervised dialogue systems adopt the BLEU score for evaluation. Ritter et al. \shortcite{ritter2011data} formulate the unsupervised learning problem as one of translating a context into a candidate response. They use a statistical machine translation (SMT) model to generate responses to various contexts using Twitter data, and show that it outperforms information retrieval baselines according to both BLEU and human evaluations. Sordoni et al.\@ \shortcite{sordoni2015aneural} extend this idea using a recurrent language model to generate responses in a context-sensitive manner. They also evaluate using BLEU, however they produce multiple ground truth responses by retrieving 15 responses from elsewhere in the corpus, using a simple bag-of-words model. Li et al.\@ \shortcite{li2015diversity} evaluate their proposed diversity-promoting objective function for neural network models using BLEU score with only a single ground truth response. A modified version of BLEU, deltaBLEU~\cite{galley2015deltableu}, which takes into account several human-evaluated ground truth responses, is shown to have a weak to moderate correlation to human judgements using Twitter dialogues. However, such human annotation is often infeasible to obtain in practice. Galley et al.\@~\shortcite{galley2015deltableu} also show that, even with several ground truth responses available, the standard BLEU metric does not correlate strongly with human judgements.

There has been significant previous work that evaluates how well automatic metrics correlate with human judgements in  in both machine translation \cite{callison2010findings,callison2011findings,bojar2014findings,graham2015accurate} and natural language generation (NLG) \cite{stent2005evaluating,cahill2009correlating,reiter2009investigation,espinosa2010further}. There has also been work criticizing the usefulness of BLEU in particular for machine translation \cite{callison2006re}. While many of the criticisms in these works apply to dialogue generation, we note that generating dialogue responses conditioned on the conversational context is in fact a more difficult problem.
%This is because most of the difficulty in automatically evaluating language generation models lies in the \textit{high entropy} of correct answers.
This is because most of the difficulty in automatically evaluating language generation models lies in the large set of correct answers.
Dialogue response generation given solely the context intuitively has a higher diversity (or entropy) than translation given text in a source language, or surface realization given some intermediate form \cite{artstein2009semi}.

\section{Evaluation Metrics}

Given a dialogue context and a proposed response, our goal is to automatically evaluate how appropriate the proposed response is to the conversation. We focus on metrics that compare it to the ground truth response of the conversation. In particular, we investigate two approaches: word based similarity metrics and word-embedding based similarity metrics.

\subsection{Word Overlap-based Metrics}

We first consider metrics that evaluate the amount of \textit{word-overlap} between the proposed response and the ground-truth response. We examine the BLEU and METEOR scores that have been used for machine translation, and the ROUGE score that has been used for automatic summarization. %\footnote{We only provide summaries of the metrics; we will add the mathematical details of all metrics using the extra page available at publication time.} %, and CIDEr which has been used for image captions. 
While these metrics have been shown to correlate with human judgements in their target domains~\cite{papineni2002bleu,lin2004rouge}, they have not been thoroughly investigated for dialogue systems.\footnote{To the best of our knowledge, only BLEU has been evaluated in the dialogue system setting quantitatively by Galley et al. \shortcite{GalleyBSJAQMGD15} on the Twitter domain. However, they carried out their experiments in a very different setting with multiple ground truth responses, which are rarely available in practice, and without providing any qualitative analysis of their results.}

%We denote the set of ground truth responses as $R$ and the set of proposed responses as $\hat{R}$. The size of both sets is $M=1$ (that is we assume that there is a single candidate ground truth response)  %(\textbf{Ryan: I don't understand what this means. Also, we should probably use a different symbol other than N, since we use that for BLEU-N.}) 
%and individual sentences are indexed using $i$ ($r_i \in R$). The $j$'th token in sentence $r_i$ is denoted $w_{ij}$.

We denote the ground truth response as $r$ (thus we assume that there is a single candidate ground truth response), and the proposed response as $\hat{r}$. The $j$'th token in the ground truth response $r$ is denoted by $w_{j}$, with $\hat{w}_j$ denoting the $j$'th token in the proposed response $\hat{r}$.

%These metrics originated for machine translation, and have been shown to correlate with human judgments in this domain~\cite{papineni2002bleu}. They have also been commonly used to evaluate end-to-end dialogue models.

\paragraph{BLEU.}
BLEU~\cite{papineni2002bleu} analyzes the co-occurrences of n-grams in the ground truth and the proposed responses. It first computes an n-gram precision for the whole dataset (we assume that there is a single candidate ground truth response per context):
$$
P_n(r,\hat{r}) = \frac{\sum_k \min(h(k,r), h(k,\hat{r}_i))}{\sum_k h(k,r_i)}
$$
where $k$ indexes all possible n-grams of length $n$ and $h(k,r)$ is the number of n-grams $k$ in $r$.\footnote{Note that the min in this equation is calculating the number of co-occurrences of n-gram $k$ between the ground truth response $r$ and the proposed response $\hat{r}$, as it computes the fewest appearances of $k$ in either response. } To avoid the drawbacks of using a precision score, namely that it favours shorter (candidate) sentences, the authors introduce a brevity penalty. BLEU-N, where $N$ is the maximum length of n-grams considered, is defined as:
%$$
%b(C,\hat{C}) = 
%\begin{cases}
				 %1\text{\,\,\,\, \,\,\,\,\,\,\,\,\,\quad \quad if $|c_i|$ \textgreater min $l_\hat{C}$,}\\
                 %e^{1-l_\hat{C}/l_C}\text{ \,\,\,\,\,\, otherwise}
%\end{cases}
%$$
%where $l_C$ is the length of a candidate sentence $c_i$, and $l_S$ is the length of the target translation. The  BLEU score is %computed by taking a etric mean of the n-gram precisions:
$$
\text{\small BLEU-N} := b(r,\hat{r}) \exp (\sum^N_{n=1} \beta_n \log P_n(r,\hat{r}))
$$
 $\beta_n$ is a weighting that is usually uniform, and $b(\cdot)$ is the brevity penalty. The most commonly used version of BLEU uses $N=4$. Modern versions of BLEU also use sentence-level smoothing, as the geometric mean often results in scores of 0 if there is no 4-gram overlap~\cite{chen2014systematic}. Note that BLEU is usually calculated at the corpus-level, and was originally designed for use with multiple reference sentences.%, and has been shown to correlate with human judgements in the translation domain when there are multiple ground truth candidates available.

%The cornerstone of the BLEU score~\cite{papineni2002bleu} is the precision measure: one measures precision by counting the number of candidate response words which also occur in the target response, and then divide by the total number of words in the generated translation. This is done for all n-grams up to length $N$ (commonly $N$=4) in the candidate response, and averaged using the etric mean to compute a modified precision score.
%$$
%p_n = \frac{\sum_{C\in{Candidates}} \sum_{n-gram\in C} %Count_{clip}(n-gram)}{\sum_{C'\in{Candidates}} \sum_{n-gram'\in C'} Count(n-gram')}
%$$

%In order to alleviate the problem of a system over-generating `reasonable' words to achieve higher precision scores, an exponential brevity penalty factor. 

\paragraph{METEOR.}
%The METEOR metric~\cite{banerjee2005meteor} was introduced to address several weaknesses in the BLEU metric. It generates an explicit alignment between words in the candidate and target sentences
%The METEOR metric \cite{banerjee2005meteor} was introduced to address several weaknesses in the BLEU metric, including the lack of recall, the lack of an explicit measure of grammaticality, and the use of geometric averaging. The latter is cited as causing BLEU scores at the sentence level to be meaningless if one of the component n-gram scores is zero. Further, BLEU doesn't conduct explicit word-to-word matching, which can lead to many incorrect matches on common filler words. 

The METEOR metric \cite{banerjee2005meteor} was introduced to address several weaknesses in BLEU. It creates an explicit alignment between the candidate and target responses. The alignment is based on exact token matching, followed by WordNet synonyms, stemmed tokens, and then paraphrases. Given a set of alignments, the METEOR score is the harmonic mean of precision and recall between the proposed and ground truth sentence.

%METEOR alleviates these issues by creating an alignment between the candidate and target responses. The aligment is based on exact token matching, followed by WordNet synonyms, stemmed tokens using the Porter stemmer, and then paraphrases. Given a set of alignments $m$, the METEOR score is the harmonic mean of precision $P_m$ and recall $R_m$ between the candidate and target sentence.
%\begin{gather} 
%Pen = \gamma (\frac{ch}{m})^\theta \\
%F_{mean} = \frac{P_m R_m}{\alpha P_m + (1 - \alpha) R_m} \\
%P_m = \frac{|m|}{\sum_k h_k(c_i)} \\
%R_m = \frac{|m|}{\sum_k h_k(s_{ij})} \\
%METEOR = (1 - Pen)F_{mean}
%\end{gather}
%The penalty term $Pen$ is based on the `chunkiness' of the resolved matches. We use the default values for the hyperparameters %$\alpha, \gamma$, and $\theta$.\footnote{See \url{http://www.cs.cmu.edu/~alavie/METEOR/}}

%It uses the Porter stemmer to take into account words with identical stems, contains a ``WN synonymy'' module that considers synonymous words, and computes a weighted F score as opposed to simply precision. This is shown to correlate more strongly with human judgements than the BLEU score in the translation domain~\cite{banerjee2005meteor}.

\paragraph{ROUGE.} ROUGE~\cite{lin2004rouge} is a set of evaluation metrics used for automatic summarization. We consider ROUGE-L, which is a F-measure based on the Longest Common Subsequence (LCS) between a candidate and target sentence. The LCS is a set of words which occur in two sentences in the same order; however, unlike n-grams the words do not have to be contiguous, i.e.\@ there can be other words in between the words of the LCS. %ROUGE-L is computed using an F-measure between the ground-truth response and the proposed response. 

\subsection{Embedding-based Metrics}

An alternative to using word-overlap based metrics is to consider the meaning of each word as defined by a \textit{word embedding}, which assigns a vector to each word. Methods such as Word2Vec~\cite{mikolov2013distributed} calculate these embeddings using distributional semantics; that is, they approximate the meaning of a word by considering how often it co-occurs with other words in the corpus.\footnote{To maintain statistical independence between the task and each performance metric, it is important that the word embeddings used are trained on corpora which do not overlap with the task corpus.} 
%Otherwise the assumptions of independent and identically distributed (i.i.d.) training and test data examples are incorrect, which could lead to spurious and potentially misleading correlations between data examples.}. 
These embedding-based metrics usually approximate sentence-level embeddings using some heuristic to combine the vectors of the individual words in the sentence.
%using word embeddings trained on an external corpus, such as Word2Vec.
The sentence-level embeddings between the candidate and target response are compared using a measure such as cosine distance. 
%This does not depend on exact word-overlap between generated and actual responses, but still allows a  quantitative comparison between responses generated by a dialogue system and the actual response of the conversation.

\paragraph{Greedy Matching.}
Greedy matching is the one embedding-based metric that does not compute sentence-level embeddings. 
%an a method for analyzing semantic similarity between sentences, which has been used for intelligent tutoring systems \cite{Rus:2012:CGO:2390384.2390403}. %It is a greedy approach to the optimization problem of matching the most similar word in sentence $T$ to a word $T'$. 
Instead, given two sequences $r$ and $\hat{r}$, each token $w \in r$ is greedily matched with a token $\hat{w} \in \hat{r}$ based on the cosine similarity of their word embeddings ($e_w$), and the total score is then averaged across all words:
\begin{gather*} 
G(r, \hat{r}) = \frac{\sum_{w \in r;} \max_{\hat{w}\in\hat{r}} cos\_sim(e_w, e_{\hat{w}}) }{|r|} \\
GM(r, \hat{r}) = \frac{G(r, \hat{r}) + G(\hat{r}, r)}{2}
\end{gather*}
This formula is asymmetric, thus we must average the greedy matching scores $G$ in each direction.
%compute it in both directions and obtain the mean score. 
This was originally introduced for intelligent tutoring systems~\cite{Rus:2012:CGO:2390384.2390403}. The greedy approach favours responses with key words that are semantically similar to those in the ground truth response. % One may argue that it is more fine-grained than the Embedding Average since it may capture multiple independent semantic elements. As before, 
%If either the ground truth response or the retrieved response does not contain any words with embeddings, their similarity is defined to be zero. 

\paragraph{Embedding Average.}
The embedding average metric calculates sentence-level embeddings using additive composition, a method for computing the meanings of phrases by averaging the vector representations of their constituent words \cite{foltz1998measurement,landauer1997solution,mitchell2008vector}. This method has been widely used in other domains, for example in textual similarity tasks \cite{DBLP:journals/corr/WietingBGL15a}. 
%In addition to the two aforementioned approaches which have been previously used, we consider a simpler model based on additive composition of embeddings. 
The embedding average, $\bar{e}$, is defined as the mean of the word embeddings of each token in a sentence $r$:
$$
\bar{e}_r = \frac{\sum_{w\in r} e_w}{|\sum_{w'\in r} e_{w'}|}.
$$
To compare a ground truth response $r$ and retrieved response $\hat{r}$, we compute the cosine similarity between their respective sentence level embeddings: $\text{EA} := \text{cos}(\bar{e}_{r},\bar{e}_{\hat{r}})$.

\paragraph{Vector Extrema.}
Another way to calculate sentence-level embeddings is using vector extrema~\cite{Forgues2014}. For each dimension of the word vectors, take the most extreme value amongst \textit{all word vectors in the sentence}, and use that value in the sentence-level embedding:
%One way to obtain sentence-level compositionality is by taking the extreme points of the word embeddings along each dimension.
$$
\small
e_{rd} = \left\{
\begin{array}{ll}
     \max_{w\in r} e_{wd}  &\text{if } e_{wd} > |\min_{w'\in r} e_{w'd}| \\
     \min_{w \in r} e_{wd}  &\text{otherwise}
\end{array}\right.
$$
where $d$ indexes the dimensions of a vector; $e_{wd}$ is the $d$'th dimensions of $e_w$ ($w$'s embedding). The min in this equation refers to the selection of the largest negative value, if it has a greater magnitude than the largest positive value.

Similarity between response vectors is again computed using cosine distance. Intuitively, this approach prioritizes informative words over common ones; words that appear in similar contexts will be close together in the vector space. Thus, common words are pulled towards the origin because they occur in various contexts, while words carrying important semantic information will lie further away. By taking the extrema along each dimension, we are thus more likely to ignore common words. %This was recently proposed by \cite{Forgues2014} in the context of bootstrapping dialogue systems with word embeddings.

% when training word embeddings using distributional semantics,
%In all three methods, if either the ground truth or proposed response does not contain any words with embeddings, their similarity is defined to be zero. This penalizes models which generate empty responses, or responses with `unknown' words, e.g.\@ words for which do not have embeddings.

%Forgues et al. \shortcite{Forgues2014} proposed the idea of taking the vector extrema, where the semantic is emphasized by taking only the extreme points and discarding the values near zero. The resulting vector extrema $E$ has the same dimension as the word embeddings $R^d$. 

%\begin{align*}
%E = extrema(T), E' = extrema(T') \\
%score(T,T') = cos\_sim( E, E' )
%\end{align*}

\section{Dialogue Response Generation Models}

%We now describe the retrieval and generative models used to train the dialogue systems considered during the evaluation of our proposed metrics.

%We now describe a variety of models that can be used to produce a response given the context of a conversation. These fall into two categories: \textit{retrieval} models, and \textit{generative} models. While we do not consider all available models, those selected cover a diverse range of end-to-end models that appear in recent literature, and provide a good sample of models for illustrating evaluation with existing metrics.

In order to determine the correlation between automatic metrics and human judgements of response quality, we obtain response from a diverse range of response generation models in the recent literature, including both \textit{retrieval} and \textit{generative} models.

\begin{table*}[!hbtp]
\footnotesize
\centering
\begin{tabular}{|l|c|c|c|c|c|c|}
\hline
  & \multicolumn{3}{|c|}{\textbf{Ubuntu Dialogue Corpus}} & \multicolumn{3}{|c|}{\textbf{Twitter Corpus}} \\ \cline{2-7}
 & Embedding & Greedy & Vector & Embedding & Greedy & Vector \\
 & Averaging & Matching & Extrema& Averaging & Matching & Extrema \\ \hline
\textsc{R-TFIDF} & 0.536 $\pm$ 0.003 & 0.370 $\pm$ 0.002 & 0.342 $\pm$ 0.002 & 0.483 $\pm$ 0.002 & 0.356 $\pm$ 0.001 & 0.340 $\pm$ 0.001  \\ 
\textsc{C-TFIDF} & 0.571 $\pm$ 0.003 & 0.373 $\pm$ 0.002 & 0.353 $\pm$ 0.002 & 0.531 $\pm$ 0.002 & 0.362 $\pm$ 0.001 & 0.353 $\pm$ 0.001 \\
\textsc{DE} & \textbf{0.650 $\pm$ 0.003} & 0.413 $\pm$ 0.002 & 0.376 $\pm$ 0.001 & \textbf{0.597 $\pm$ 0.002} & 0.384 $\pm$ 0.001 & 0.365 $\pm$ 0.001 \\ \hline
\textsc{LSTM} & 0.130 $\pm$ 0.003 & 0.097 $\pm$ 0.003 & 0.089 $\pm$ 0.002 &0.593 $\pm$ 0.002& \textbf{0.439 $\pm$ 0.002}& \textbf{0.420 $\pm$ 0.002} \\
\textsc{HRED} & 0.580 $\pm$ 0.003 & \textbf{0.418 $\pm$ 0.003} & \textbf{0.384 $\pm$ 0.002} & \textbf{0.599 $\pm$ 0.002} & \textbf{0.439 $\pm$ 0.002}& \textbf{0.422 $\pm$ 0.002} \\ \hline
\end{tabular}
\caption{\label{results}Models evaluated using the vector-based evaluation metrics, with 95\% confidence intervals.}
\end{table*}

\subsection{Retrieval Models}

Ranking or retrieval models for dialogue systems are typically evaluated based on whether they can retrieve the correct response from a corpus of pre-defined responses, which includes the ground truth response to the conversation~\cite{schatzmann2005quantitative}. Such systems can be evaluated using recall or precision metrics. However, when deployed in a real setting these models will not have access to the correct response given an unseen conversation. Thus, in the results presented below we \textit{remove} one occurrence of the ground-truth response from the corpus and ask the model to retrieve the most appropriate response from the remaining utterances. Note that this does not mean the correct response will not appear in the corpus at all; in particular, if there exists another context in the dataset with an identical ground-truth response, this will be available for selection by the model.

%TO ADD back in
%More specifically, each retrieval model specifies a scoring function $S$ between the context $c$ and a candidate response in the corpus. The model retrieves a response $r^*$ from the corpus such that:
%$$
%r^* =  \underset{r}{\operatorname{argmax}} [ S(c, r) ], \quad r \in R,
%$$
%where $R$ is the set of all responses being considered. These can be from either the training or test set, but \textit{cannot} include the actual response of the conversation. 

%\textbf{Julian: You need to more clear about this? Are the models searching through the training set only? Are they searching through the test set minus the correct example? Or are they searching through both?}
We then evaluate each model by comparing the retrieved response to the ground truth response of the conversation. %To our knowledge, this is a novel interpretation of the evaluation of retrieval-based dialogue models. 
This  closely imitates real-life deployment of these models, as it tests the ability of the model to generalize to unseen contexts.

%The response retrieval task is an extension of next utterance classification. If the dialogue agent is able to retrieve from a set of \textit{all} possible responses, we can even argue that it becomes a generative model. Using the same Dual-Encoder model, trained on next utterance classification, we retrieve the best response $r$ for a given context $c$ as follows:

%$$r^* = \underset{r}{\operatorname{argmax}}  \;\; \sigma(c^TMr+b) \;\;\; \forall r \in R$$
%where $R$ contains the entire set of candidate response vectors, computed by the RNN. 

\paragraph{TF-IDF.}

We consider a simple Term Frequency - Inverse Document Frequency (TF-IDF) retrieval model~\cite{lowe2015ubuntu}. TF-IDF is a statistic that intends to capture how important a given word is to some document, which is calculated as:
$
\textmd{tfidf}(w,c,C) = f(w,c) \times \log \frac{N}{|\{c \in C : w \in c\}|},
$
where $C$ is the set of all contexts in the corpus, $f(w,c)$ indicates the number of times word $w$ appeared in context $c$, $N$ is the total number of dialogues, and the denominator represents the number of dialogues in which the word $w$  appears. 

In order to apply TF-IDF as a retrieval model for dialogue, we first compute the TF-IDF vectors for each context and response in the corpus. We then return the response with the largest cosine similarity in the corpus, either between the input context and corpus contexts (C-TFIDF), or between the input context and corpus responses (R-TFIDF).

\paragraph{Dual Encoder.}

Next we consider the recurrent neural network (RNN) based architecture called the Dual Encoder (\textsc{DE}) model~\cite{lowe2015ubuntu}. The DE model consists of two RNNs which respectively compute the vector representation of an input context and response, $c,r \in \mathbb{R}^n$. The model then calculates the probability that the given response is the ground truth response given the context, by taking a weighted dot product:
$p(r\text{ is correct}|c,r,M) = \sigma(c^T M r + b)$
where $M$ is a matrix of learned parameters and $b$ is a bias. 
%Intuitively, this corresponds to projecting each candidate response $r$ into some `context space' via the matrix product $Mr$, and measuring the similarity between the generated context and the context of the conversation with the dot product.
The model is trained using negative sampling to minimize the cross-entropy error of all (context, response) pairs. To our knowledge, our application of neural network models to large-scale retrieval in dialogue systems is novel.

\begin{table*}[!hbtp]
\small
\centering
\begin{tabular}{|l|c c|c c|c c| c c|}
\hline
 & \multicolumn{4}{|c|}{\textbf{Twitter}} & \multicolumn{4}{|c|}{\textbf{Ubuntu}} \\ \hline

Metric & \textbf{Spearman }& p-value & \textbf{Pearson} & p-value & \textbf{Spearman }& p-value & \textbf{Pearson} & p-value \\ \hline
Greedy  & 0.2119 & 0.034 & 0.1994 & 0.047 & 0.05276 & 0.6 & 0.02049 & 0.84 \\
Average & 0.2259 & 0.024 & 0.1971 & 0.049 & -0.1387 & 0.17 & -0.1631 & 0.10 \\
Extrema & 0.2103 & 0.036 & 0.1842 & 0.067 & 0.09243 & 0.36 & -0.002903 & 0.98 \\
METEOR  & 0.1887 & 0.06 & 0.1927 & 0.055 & 0.06314 & 0.53 & 0.1419 & 0.16 \\
BLEU-1   & 0.1665 & 0.098 & 0.1288 & 0.2 & -0.02552 & 0.8 & 0.01929 & 0.85 \\
BLEU-2   & 0.3576 & $<$ 0.01 & 0.3874 & $<$ 0.01 & 0.03819 & 0.71 & 0.0586 & 0.56 \\
BLEU-3   & 0.3423 & $<$ 0.01 & 0.1443 & 0.15 & 0.0878 & 0.38 & 0.1116 & 0.27 \\
BLEU-4   & 0.3417 & $<$ 0.01 & 0.1392 & 0.17 & 0.1218 & 0.23 & 0.1132 & 0.26 \\
ROUGE   & 0.1235 & 0.22 & 0.09714 & 0.34 & 0.05405 & 0.5933 & 0.06401 & 0.53 \\ \hline
Human   & 0.9476 & $<$ 0.01 & 1.0  & 0.0 & 0.9550 & $<$ 0.01 & 1.0 & 0.0 \\
%CIDEr   & -      & -      & -       & -      & -        & -    & -         & -    \\
\hline
\end{tabular}
\caption{\label{tab:correlation}  Correlation between each metric and human judgements for each response. Correlations shown in the human row result from randomly dividing human judges into two groups.}
\end{table*}
% Unmodified	Without Stopwords/Punctuation
%Spearman	Pearson	Spearman	Pearson
%BLEU1	0.1665	0.1288	0.1421	0.1521
%BLEU2	0.3576	0.3874	0.1982	0.1979
%BLEU3	0.3423	0.1443	0.2315	0.2435
%BLEU4	0.3417	0.1392	0.2253	0.264

\begin{table}[!hbtp]
    \footnotesize
    \centering
    \begin{tabular}{|l|c c |c c|}
        \hline
         & \textbf{Spearman} & p-value & \textbf{Pearson} & p-value \\ \hline
        BLEU-1 & 0.1580 & 0.12 & 0.2074 & 0.038 \\ \hline
        BLEU-2 & 0.2030 & 0.043 & 0.1300 & 0.20\\ \hline
%        BLEU3 & 0.2315 & 0.020 & 0.2435 & 0.015 \\ \hline
%        BLEU4 & 0.2253 & 0.024 & 0.2640 & $<$ 0.01 \\ \hline
    \end{tabular}
    \caption{\label{tab:bleu_correlation} Correlation between BLEU metric and human judgements after removing stopwords and punctuation for the Twitter dataset.}
\end{table}

\begin{table}[!hbtp]
    \footnotesize
    \centering
    \begin{tabular}{|l|c c c|}
        \hline
        & \multicolumn{2}{c}{Mean score}  &  \\
         &$\Delta w <= 6$ & $\Delta w >=6$ & p-value \\
         &(n=47) & (n=53) & \\ \hline
        BLEU-1 & 0.1724 & 0.1009 & $<$ 0.01 \\ \hline
        BLEU-2 & 0.0744 & 0.04176 & $<$ 0.01 \\ \hline
        Average & 0.6587 & 0.6246 & 0.25 \\ \hline
        METEOR & 0.2386 & 0.2073 & $<$ 0.01 \\ \hline
        Human & 2.66 & 2.57 & 0.73 \\ \hline
    \end{tabular}
    \caption{\label{tab:length_effect} Effect of differences in response length for the Twitter dataset, $\Delta w$ = absolute difference in \#words between a ground truth response and proposed response}
\end{table}

\subsection{Generative Models}

In addition to retrieval models, we also consider generative models. In this context, we refer to a model as generative if it is able to generate entirely new sentences that are unseen in the training set. %The most popular such models in the recent literature use a recurrent network to generate tokens one at a time.

\paragraph{LSTM language model.}

The baseline model is an LSTM language model~\cite{hochreiter97lstm} trained to predict the next word in the (context, response) pair. During test time, the model is given a context, encodes it with the LSTM and generates a response using a greedy beam search procedure \cite{graves2013generating}. 
%During test time, the model is given a context, encodes it with the LSTM and generates a response using a greedy beam search procedure \cite{graves2013generating}.  %Note that this is similar to the Encoder-Decoder architecture from Cho et al. \shortcite{cho2014learning}, where the encoder and decoder have tied weights. %TODO: confirm this

%The baseline model is a simple LSTM language model (LSTM LM)~\cite{hochreiter97lstm}, a neural network model which is used to predict the next word in a sequence of words \cite{mikolov2010recurrent}. The model observes the dialogue word-by-word and updates it hidden state $h_t$ at time step (word index) $t$. Given a hidden state $h_t$ the model outputs a probability distribution over all words in the vocabulary. Formally, the model computes the hidden state
%$h_t = f(h_{t-1}, y_{t-1}).$
%The conditional distribution over each output symbol is computed in a similar manner:
%$$
%P(y_t|y_{t-1}, ..., y_1,) = g(h_t, y_{t-1}),
%$$
%for activation functions $f$ and $g$. 

%The model is trained in an end-to-end fashion by gradient descent to maximize the conditional log-likelihood of input-output pairs from the training set, %$(y_n)$:
%$$
%\max_\theta \frac{1}{N} \sum^N_{n=1} \log p_\theta (y_n),
%$$
%where $\theta$ are the parameters of the model.
%Thus, the model learns a probability distribution over all output sequences, 
%$p(y_1, ..., y_{T})$. %A diagram of a typical encoder-decoder model can be seen in Figure \_\_.

\paragraph{HRED.}

Finally we consider the Hierarchical Recurrent Encoder-Decoder (\textsc{HRED}) \cite{2015arXiv150704808S}.
In the traditional Encoder-Decoder framework, all utterances in the context are concatenated together before encoding. Thus, information from previous utterances is far outweighed by the most recent utterance.
The HRED model uses a \textit{hierarchy} of encoders; each utterance in the context passes through an `utterance-level' encoder, and the output of these encoders is passed through another `context-level' encoder, which enables the handling of longer-term dependencies. 
%. This enables the handling of longer-term dependencies compared to a conventional Encoder-Decoder. %The decoding is done with another LSTM, using beam search.

\subsection{Conclusions from an Incomplete Analysis}

When evaluation metrics are not explicitly correlated to human judgement, it is possible to draw misleading conclusions by examining how the metrics rate different models. %Even if multiple metrics show that one model significantly improves over other models across multiple domains, there is no guarantee that that humans will arrive at the same conclusion.
To illustrate this point, we compare the performance of selected models according to the embedding metrics on two different domains: the Ubuntu Dialogue Corpus \cite{lowe2015ubuntu}, which contains technical vocabulary and where conversations are often oriented towards solving a particular problem, and a non-technical Twitter corpus collected following the procedure of Ritter et al.\@ \shortcite{Ritter:2010:UMT:1857999.1858019}. %, where the dialogues cover a diverse set of topics often without any particular goal. %We also evaluate the performance of the retrieval models on a Twitter corpus, collected following the procedure outlined by Ritter et al.\@ \shortcite{Ritter:2010:UMT:1857999.1858019}.
We consider these two datasets since they cover contrasting dialogue domains, i.e.\@ technical help vs casual chit-chat, and because they are amongst the largest publicly available corpora, making them good candidates for building data-driven dialogue systems. 

Results on the proposed embedding metrics are shown in Table \ref{results}. For the retrieval models, we observe that the \textsc{DE} model significantly outperforms both TFIDF baselines on all metrics across both datasets. %We also observed this qualitatively, as the DE model is much better at capturing the semantics of the context.
Further, the \textsc{HRED} model significantly outperforms the basic LSTM generative model in both domains, and appears to be of similar strength as the DE model. 
Based on these results, one might be tempted to conclude that there is some information being captured by these metrics, that significantly differentiates models of different quality. However, as we show in the next section, the embedding-based metrics correlate only weakly with human judgements on the Twitter corpus, and not at all on the Ubuntu Dialogue Corpus. This demonstrates that metrics that have not been specifically correlated with human judgements on a new task should not be used to evaluate that task.

\begin{figure*}
\centering

\begin{subfigure}[b]{1.0\textwidth}
            \centering
            \includegraphics[width=0.3\linewidth]{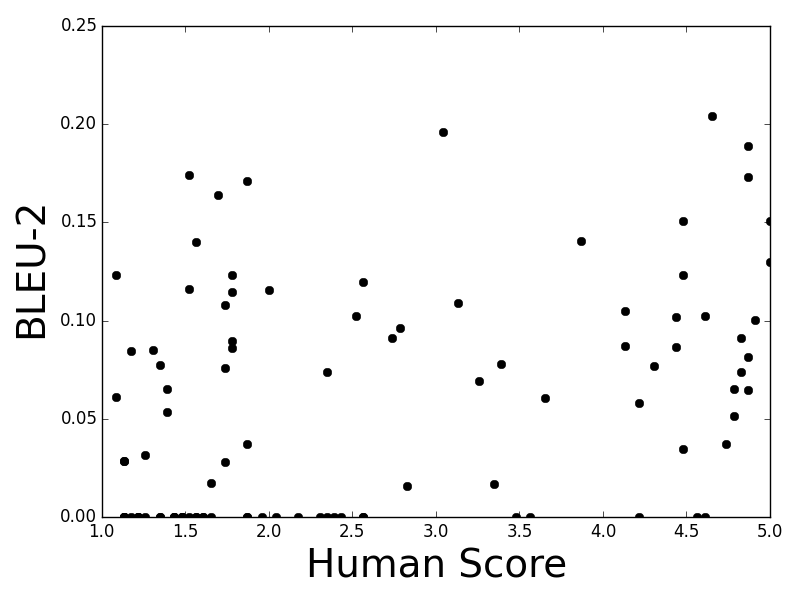}
             \includegraphics[width=0.3\linewidth]{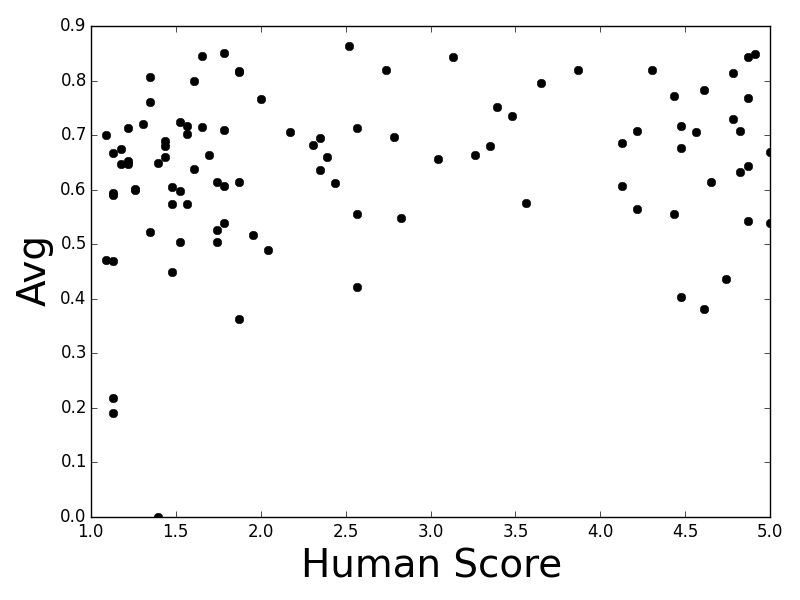}
            \includegraphics[width=0.3\linewidth]{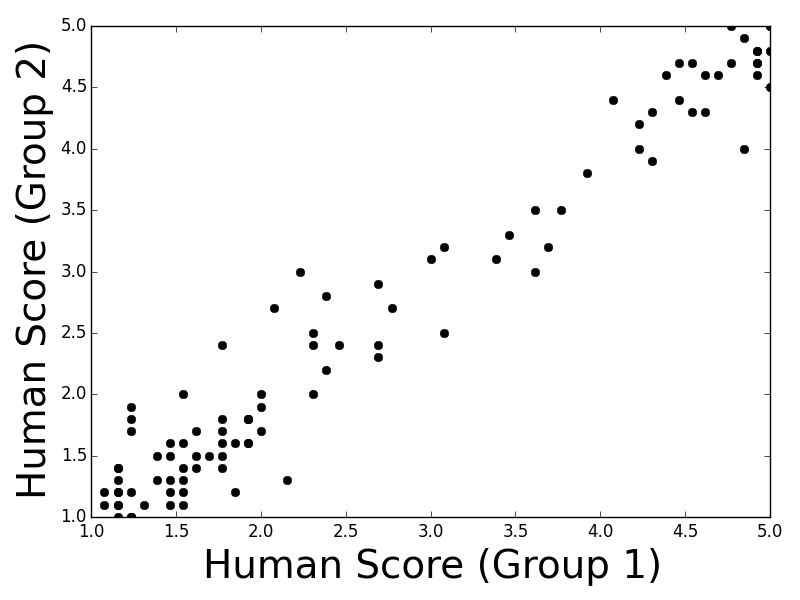}
            \caption{Twitter}
\end{subfigure}\\
\begin{subfigure}[b]{1.0\textwidth}
            \centering
            \includegraphics[width=0.3\linewidth]{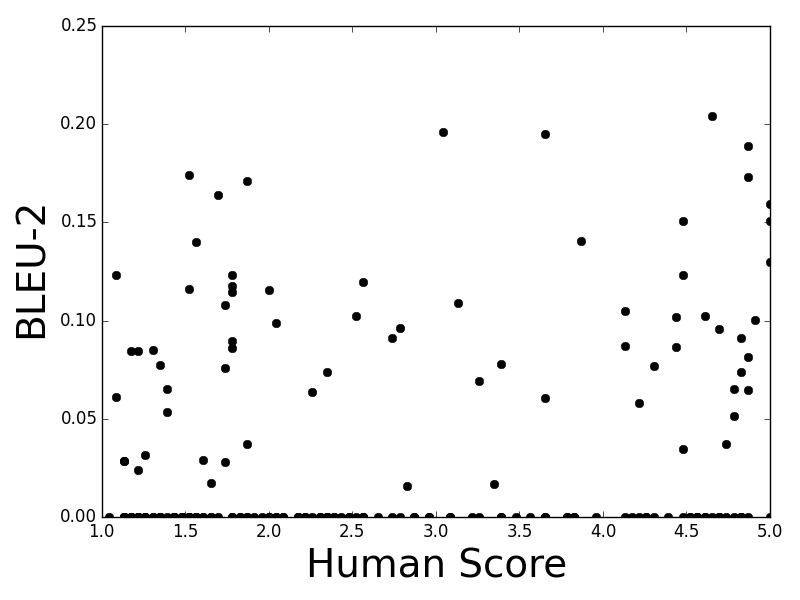}
            \includegraphics[width=0.3\linewidth]{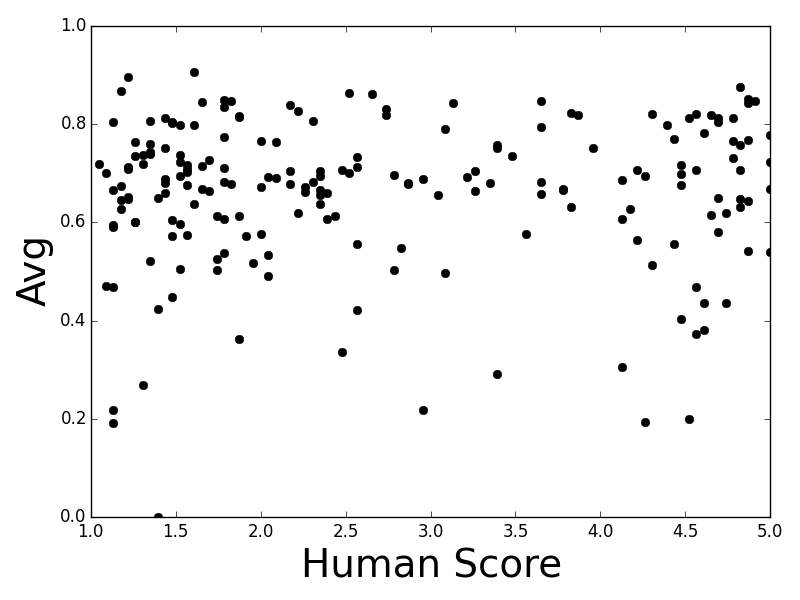}
            \includegraphics[width=0.3\linewidth]{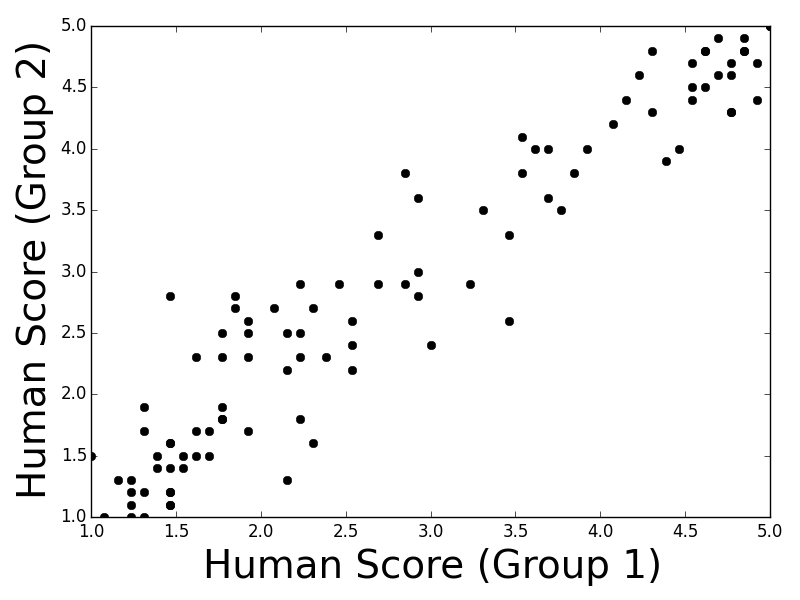}
            \caption{Ubuntu}
\end{subfigure}\\

\caption{\label{fig:scatter} Scatter plots showing the correlation between metrics and human judgements on the Twitter corpus (a) and Ubuntu Dialogue Corpus (b). The plots represent BLEU-2 (left), embedding average (center), and correlation between two randomly selected halves of human respondents (right).}
\end{figure*}

\section{Human Correlation Analysis}\label{sec:human_correlation_analysis}

\paragraph{Data Collection.}

We conducted a human survey to determine the correlation between human judgements on the quality of responses, and the score assigned by each metric. We aimed to follow the procedure for the evaluation of BLEU~\cite{papineni2002bleu}. 25 volunteers from the Computer Science department at the author's institution were given a context and one proposed response, and were asked to judge the response quality on a scale of 1 to 5.\footnote{Studies asking humans to evaluate text often rate different aspects separately, such as `adequacy', `fluency' and `informativeness' of the text \cite{hovy199toward,papineni2002corpus} Our evaluation focuses on adequacy. We did not consider fluency because 4 out of the 5 proposed responses to each context were generated by a human. We did not consider informativeness because in the domains considered, it is not necessarily important (in Twitter), or else it seems to correlate highly with adequacy (in Ubuntu).}; a 1 indicates that the response is not appropriate or sensible given the context, and a 5 indicates that the response is very reasonable.
Out of the 25 respondents, 23 had Cohen's kappa scores $\kappa > 0.2$ w.r.t.\@ the other respondents, which is a standard measure for inter-rater agreement~\cite{cohen1968weighted}. The 2 respondents with $\kappa<0.2$, indicating slight agreement, were excluded from the analysis below. The median $\kappa$ score was approximately 0.55, roughly indicating moderate to strong annotator agreement. %Further details are given in the Supplemental Materials.

Each volunteer was given 100 questions per dataset. These questions correspond to 20 unique contexts, with 5 different responses: one utterance randomly drawn from elsewhere in the test set, the response selected from each of the TF-IDF, DE, and HRED models, and a response written by a human annotator. These were chosen as they cover the range of qualities almost uniformly (see Figure \ref{fig:scatter}). %The questions were randomly permuted within each dataset during the experiments.

\begin{figure*}[!htbp]
\footnotesize
\centering
\begin{tabular}{l}
\hline
\textbf{Context of Conversation} \\ %\hline
 A: dearest! question. how many thousands of people\\
 can panaad occupy?\\
 B: @user panaad has $<$number$>$ k seat capacity while rizal \\
 has  $<$number$>$ k thats why they choose rizal i think . \\ \hline
\textbf{Ground Truth Response}\\ %\hline
 A: now i know about the siting capacity . thanks for the \\
 info @user great evening.\\ \hline
\textbf{Proposed Response}\\ %\hline
 A: @user makes sense. thanks!\\ \\ \hline
\end{tabular}
\begin{tabular}{l}
 \hline
\textbf{Context of Conversation} \\ %\hline
A: never felt more sad than i am now\\
 B: @user aww why ?\\
 A: @user @user its a long story ! sure you wanna know \\
 it ? bahaha and thanks for caring btw $<$heart$>$ \\ \hline
\textbf{Ground Truth Response}\\ %\hline
 A: @user i don 't mind to hear it i 've got all day and \\
 youre welcome $<$number$>$\\ \hline
\textbf{Proposed Response}\\ %\hline
 A: @user i know , i 'm just so happy for you ! ! ! ! ! ! ! !\\
 ! ! ! ! ! ! ! ! ! ! ! ! ! ! ! ! ! \\\hline
\end{tabular}
\caption{\label{fig:qual}Examples where the metrics rated the response poorly and humans rated it highly (left), and the converse (right). Both responses are given near-zero score by BLEU-N for N$>1$. While no metric will perform perfectly on all examples, we present these examples to provide intuition on how example-level errors become aggregated into poor correlation to human judgements at the corpus-level.}
\end{figure*}

\paragraph{Survey Results.}

We present correlation results between the human judgements and each metric in Table \ref{tab:correlation}. We compute the Pearson correlation, which estimates linear correlation, and Spearman correlation, which estimates any monotonic correlation. 

The first observation is that in both domains the BLEU-4 score, which has previously been used to evaluate unsupervised dialogue systems, shows very weak if any correlation with human judgement. In fact we found that the BLEU-3 and BLEU-4 scores were near-zero for a majority of response pairs; for BLEU-4, only four examples had a score $>10^{-9}$. Despite this, they still correlate with human judgements on the Twitter Corpus at a rate similar to BLEU-2. This is because of the smoothing constant, which gives a tiny weight to unigrams and bigrams despite the absence of higher-order n-grams. BLEU-3 and BLEU-4 behave as a scaled, noisy version of BLEU-2; thus, if one is to evaluate dialogue responses with BLEU, we recommend the choice of $N=2$ over $N=3$ or 4. Note that using a test corpus larger than the size reported in this paper may lead to stronger correlations for BLEU-3 and BLEU-4, due to a higher number of non-zero scores.
%Thus, these metrics simply act as re-scaled BLEU-2.

It is interesting to note that, while some of the embedding metrics and BLEU show small positive correlation in the non-technical Twitter domain, 
%but METEOR and ROUGE fail to show any correlation in that domain.
there is no metric that significantly correlates with humans on the Ubuntu Dialogue Corpus.
This is likely because the correct Ubuntu responses contain specific technical words that are less likely to be produced by our models.
Further, it is possible that responses in the Ubuntu Dialogue Corpus have intrinsically higher variability (or entropy) than Twitter when conditioned on the context, making the evaluation problem significantly more difficult.

Figure \ref{fig:scatter} illustrates the relationship between metrics and human judgements. We include only the best performing metric using word-overlaps, i.e.\@ the BLEU-2 score (left), and the best performing metric using word embeddings, i.e.\@ the vector average (center). These plots show how weak the correlation is: in both cases, they appear to be random noise. It seems as though the BLEU score obtains a positive correlation because of the large number of responses that are given a score of 0 (bottom left corner of the first plot).
This is in stark contrast to the inter-rater agreement, which is plotted between two randomly sampled halves of the raters (right-most plots). 
%We also observed this for BLEU3 on Twitter: although all except for 4 of the responses were given scores of zero according to this metric, there is still a significant correlation as measured using the Spearman coefficient. On the other hand, human scores correlate very strongly with each other.
We also calculated the BLEU scores after removing stopwords and punctuation from the responses. As shown in Table \ref{tab:bleu_correlation}, this weakens the correlation with human judgements for BLEU-2 compared to the values in Table \ref{tab:correlation}, and suggests that BLEU is sensitive to factors that do not change the semantics of the response.

Finally, we examined the effect of response length on the metrics, by considering changes in scores when the ground truth and proposed response had a large difference in word counts. 
%for the cases where the response length differ in terms of words, between the ground truth response and the proposed response. 
Table~\ref{tab:bleu_correlation} shows that BLEU and METEOR are particularly sensitive to this aspect, compared to the Embedding Average metric and human judgement.

\paragraph{Qualitative Analysis.}

In order to determine specifically why the metrics fail, we examine qualitative samples where there is a disagreement between the metrics and human rating. Although these only show inconsistencies at the example-level, they provide some intuition as to why the metrics don't correlate with human judgements at the corpus-level. We present in Figure \ref{fig:qual} two examples where all of the embedding-based metrics and BLEU-1 score the proposed response significantly differently than the humans.

The left of Figure \ref{fig:qual} shows an example where the embedding-based metrics score the proposed response lowly, while humans rate it highly. It is clear from the context that the proposed response is reasonable -- indeed both responses intend to express gratitude. However, the proposed response has a different wording than the ground truth response, and therefore the metrics are unable to separate the salient words from the rest. This suggests that the embedding-based metrics would benefit from a weighting of word saliency.

The right of the figure shows the reverse scenario: the embedding-based metrics score the proposed response highly, while humans do not. This is most likely due to the frequently occurring `i' token, and the fact that `happy' and `welcome' may be close together in the embedding space. However, from a human perspective there is a significant semantic difference between the responses as they pertain to the context. Metrics that take into account the context may be required in order to differentiate these responses. Note that in both responses in Figure \ref{fig:qual}, there are no overlapping n-grams greater than unigrams between the ground truth and proposed responses; thus, all of BLEU-2,3,4 would assign a score near 0 to the response.

%BLEU-2, BLEU-3, and BLEU-4

\section{Discussion}

%We introduce three word embedding-based metrics for use in evaluating dialogue systems in an unsupervised manner, without requiring task completion labels or simulated users.
%When using these metrics to evaluate retrieval and generative dialogue models, we observe that these models are comparable in their ability to generate semantically meaningful responses.
%While we believe these metrics are useful to compare sentences based on semantic content (rather than word alignment), they are by no means sufficient. We strongly advocate for automatically evaluating dialogue systems with as many relevant metrics as possible. Notably, while our metrics can evaluate the semantic content of generated responses, they are unable to evaluate coherence of responses; this is an important area for future work. 
%Further analysis is also required to determine which of these metrics correlates most strongly with human judgement. 

We have shown that many metrics commonly used in the literature for evaluating unsupervised dialogue systems do not correlate strongly with human judgement. Here we elaborate on important issues arising from our analysis.

\paragraph{Constrained tasks.}
Our analysis focuses on relatively unconstrained domains.
Other work, which separates the dialogue system into a dialogue planner and a natural language generation component for applications in constrained domains, may find stronger correlations with the BLEU metric.
For example, Wen et al.\@ \shortcite{wen2015semantically} propose a model to map from dialogue acts to natural language sentences and use BLEU to evaluate the quality of the generated sentences.
Since the mapping from dialogue acts to natural language sentences has lower diversity and is more similar to the machine translation task, it seems likely that BLEU will correlate better with human judgements.
However, an empirical investigation is still necessary to justify this.

%\paragraph{Modular systems}
%In our work we considered only models and humans that generate a response in an end-to-end manner, that is directly from an input context. Other work has focused on \textit{modular} dialogue systems that contain a separate natural language generation component, and use BLEU to evaluate generated responses in this domain~\cite{wen2015semantically}. Since these models generate utterances from dialogue acts, which are significantly more constrained, rather than directly from the context, it is plausible that BLEU does correlate with human judgement, as it is more closely related to the original translation problem.

\paragraph{Incorporating multiple responses.}
Our correlation results assume that only one ground truth response is available given each context. Indeed, this is the common setting in most of the recent literature on training end-to-end conversation models. There has been some work on using a larger set of automatically retrieved plausible responses when evaluating with BLEU~\cite{galley2015deltableu}. However, there is no standard method for doing this in the literature. Future work should examine how retrieving additional responses affects the correlation with word-overlap metrics.

\paragraph{Searching for suitable metrics.}
While we provide evidence against existing metrics, we do not yet provide good alternatives for unsupervised evaluation. Despite the poor performance of the word embedding-based metrics in this survey, we believe that metrics based on distributed sentence representations hold the most promise for the future. 
This is because word-overlap metrics will simply require too many ground-truth responses to find a significant match for a reasonable response, due to the high diversity of dialogue responses. 
%Given the high entropy nature of dialogue responses, it is unlikely that it will ever be feasible to collect enough ground-truth responses for each context to 
As a simple example, the skip-thought vectors of Kiros et al. \shortcite{kiros2015skip} could be considered.
Since the embedding-based metrics in this paper only consist of basic averages of vectors obtained through distributional semantics, they are insufficiently complex for modeling sentence-level compositionality in dialogue. Instead, these metrics can be interpreted as calculating the \textit{topicality} of a proposed response (i.e.\@ how on-topic the proposed response is, compared to the ground-truth). 

All of the metrics considered in this paper directly compare a proposed response to the ground-truth, without considering the context of the conversation. %This could be problematic, as the appropriateness of the next utterance depends much more strongly on the context than on other suitable utterances.
%Thus, metrics that take into account the context could also be considered. 
However, metrics that take into account the context could also be considered. 
Such metrics could come in the form of an \textit{evaluation model} that is learned from data.
This model could be either a discriminative model that attempts to distinguish between model and human responses, or a model that uses data collected from the human survey in order to provide human-like scores to proposed responses. 
%We are currently investigating these and other approaches, in the search for accurate automated metrics for evaluating dialogue systems.
Finally, we must consider the hypothesis that learning such models from data is no easier than solving the problem of dialogue response generation.
If this hypothesis is true,  we must concede and always use human evaluations together with metrics that only roughly approximate human judgements.
%In this case, we must concede and use metrics that only roughly approximate human judgements, along with human evaluations.

\bibliography{emnlp2016}
\bibliographystyle{emnlp2016}

\newpage

\section*{Appendix}

\subsection*{Distribution of kappa scores}

%Here we provide more details on the human experiments in the paper.
Table \ref{tab:kappa} shows the full distribution over $\kappa$ scores for each pair of human annotators. It is apparent that most of the scores (88.9\%) are over 0.4, indicating a moderate agreement. This suggests that the task was reasonable and well understood by the annotators.

\begin{table}[h]
    \centering
    \begin{tabular}{c|c|c}
         $\kappa$ & \# pairs & \% pairs  \\ \hline
         $>0.2$ & 253/253 & 100\% \\
         $>0.3$ & 251/253 & 99.2\% \\
         $>0.4$ & 225/253 & 88.9\% \\
         $>0.5$ & 162/253 & 64.0\% \\
         $>0.6$ & 50/253 & 19.8\% \\
         $>0.7$ & 3/253 & 1.2\% \\
         $>0.8$ & 0/253 & 0\% \\
    \end{tabular}
    \caption{Distribution of pairwise $\kappa$ scores between each pair of human annotators, other than the annotators that were discarded due to low scores.}
    \label{tab:kappa}
\end{table}

\subsection*{Full scatter plots}

We present the scatterplots for all of the metrics consider and their correlation with human judgement, in Figures 3-7 below. As previously emphasized, there is very little correlation for any of the metrics, and the BLEU-3 and BLEU-4 scores are often close to zero.

\begin{figure*}
\centering
\includegraphics[width=0.45\linewidth]{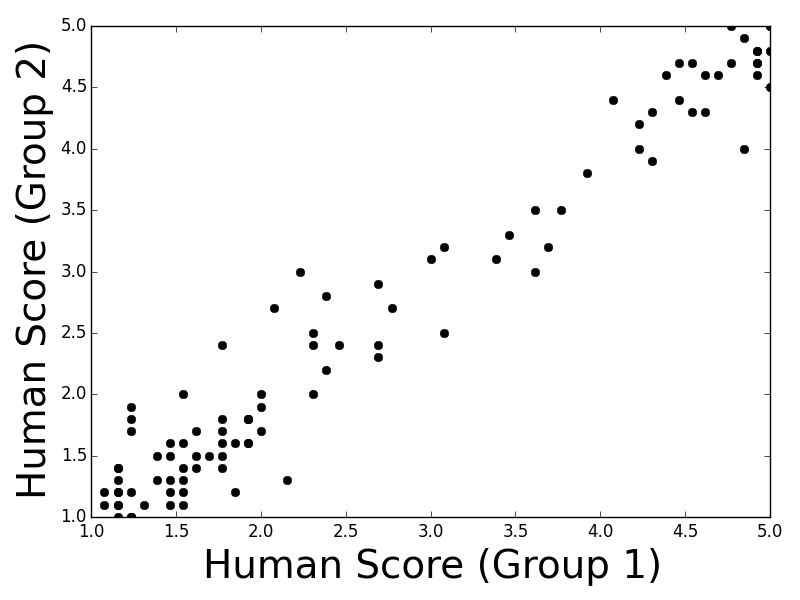}
\includegraphics[width=0.45\linewidth]{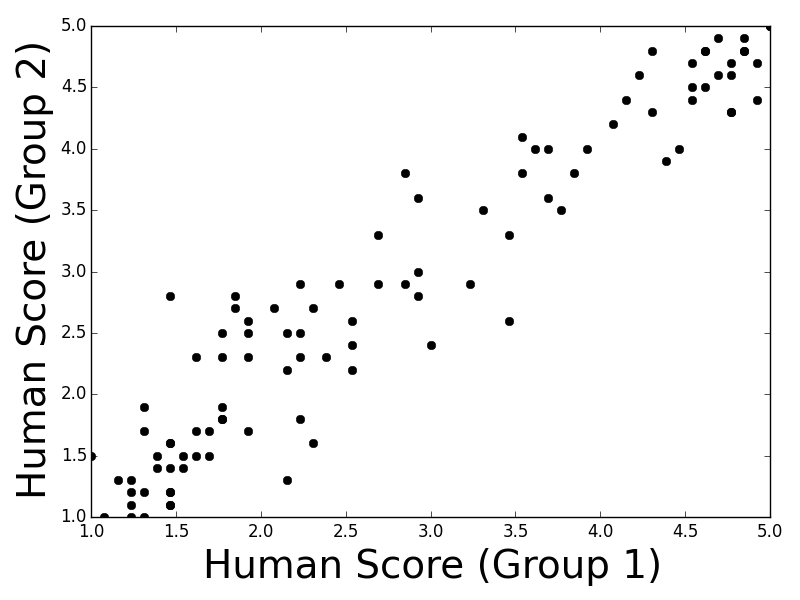}
\caption{\label{fig:scatterhum} Scatter plots showing the correlation between two randomly chosen groups of human volunteers on the Twitter corpus (left) and Ubuntu Dialogue Corpus (right).}
\end{figure*}

\begin{figure*}
\centering

\begin{subfigure}[b]{1.0\textwidth}
            \centering
            \includegraphics[width=0.48\linewidth]{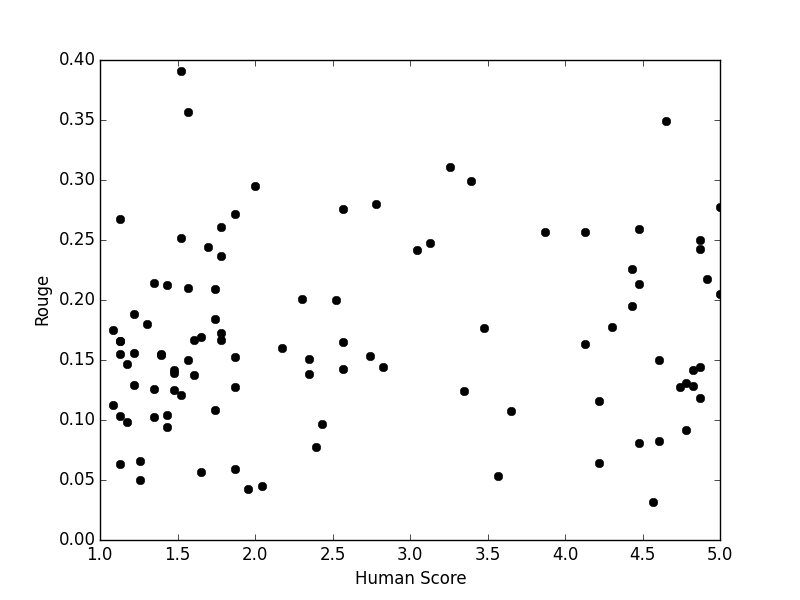}
            \includegraphics[width=0.48\linewidth]{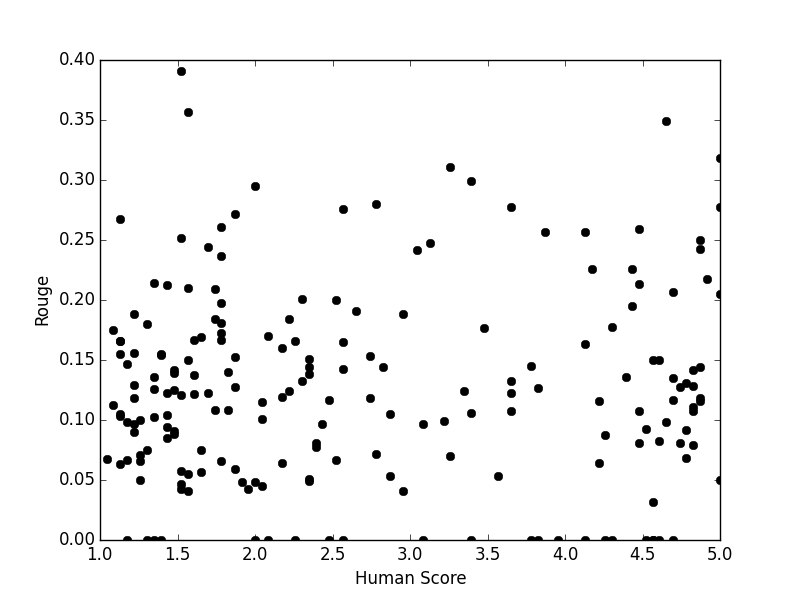}
            \caption{ROUGE}
\end{subfigure}\\
\begin{subfigure}[b]{1.0\textwidth}
            \centering
            \includegraphics[width=0.48\linewidth]{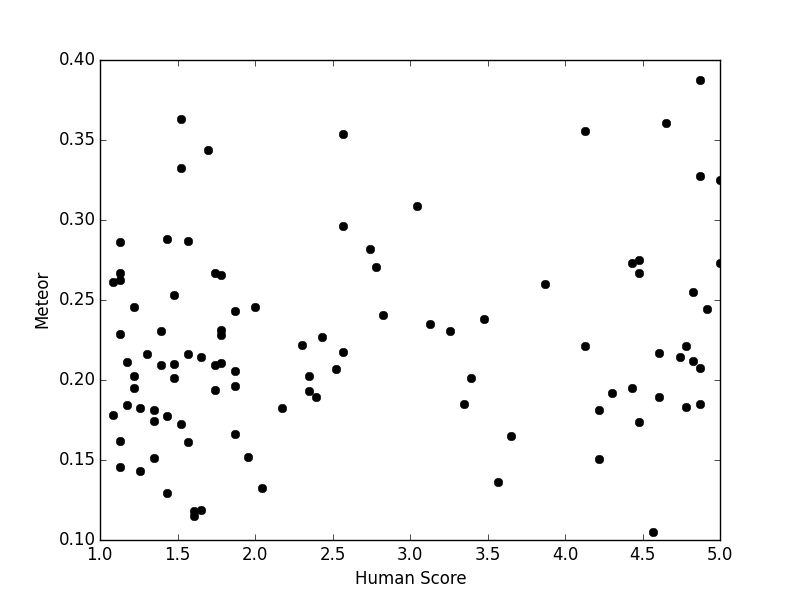}
            \includegraphics[width=0.48\linewidth]{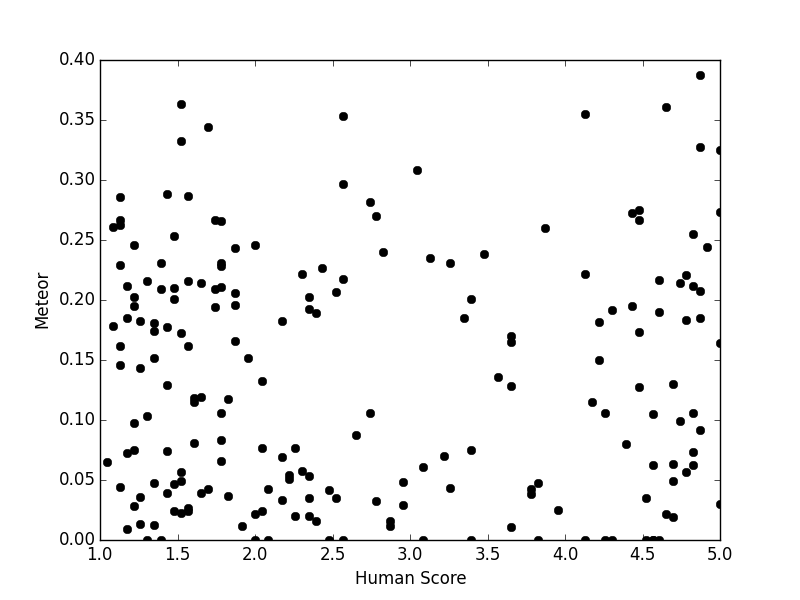}
            \caption{METEOR}
\end{subfigure}\\

\caption{\label{fig:scatterrouge} Scatter plots showing the correlation between metrics and human judgement on the Twitter corpus (left) and Ubuntu Dialogue Corpus (right). The plots represent ROUGE (a) and METEOR (b).}
\end{figure*}

\begin{figure*}
\centering

\begin{subfigure}[b]{1.0\textwidth}
            \centering
            \includegraphics[width=0.47\linewidth]{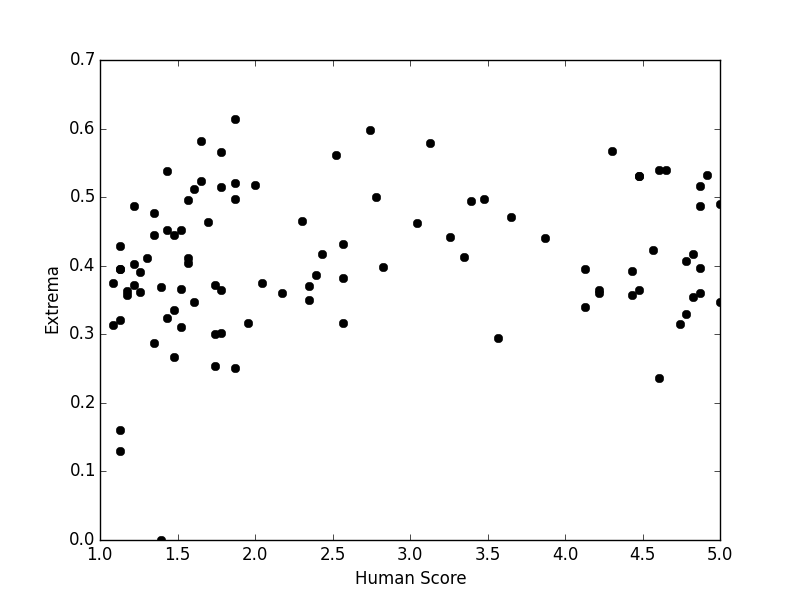}
            \includegraphics[width=0.47\linewidth]{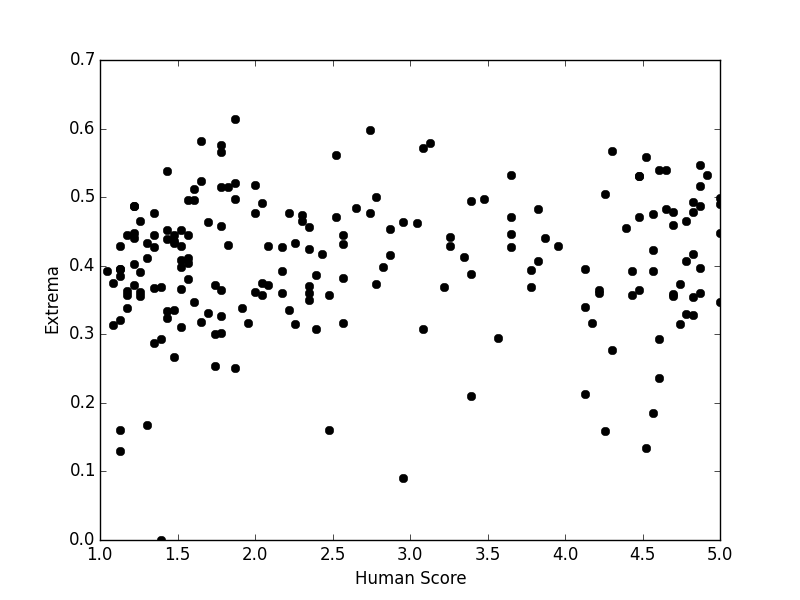}
            \caption{Vector Extrema}
\end{subfigure}\\
\begin{subfigure}[b]{1.0\textwidth}
            \centering
            \includegraphics[width=0.47\linewidth]{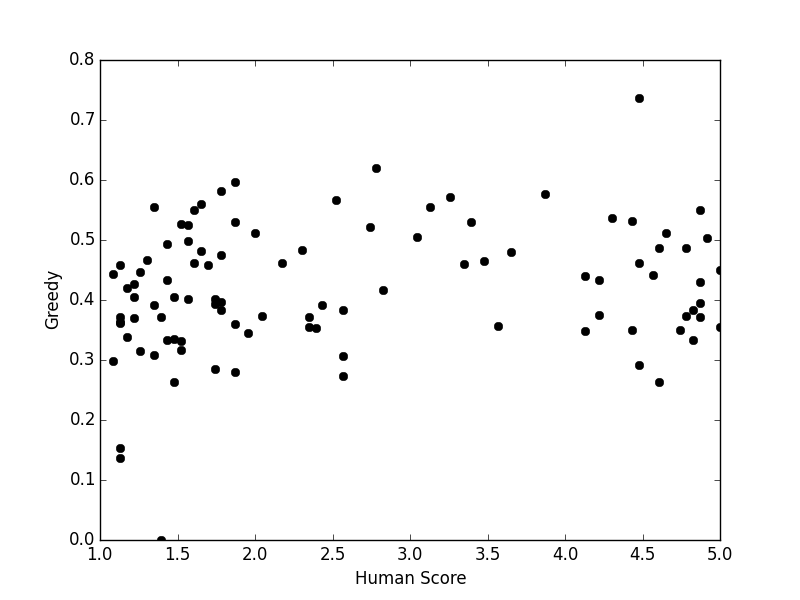}
            \includegraphics[width=0.47\linewidth]{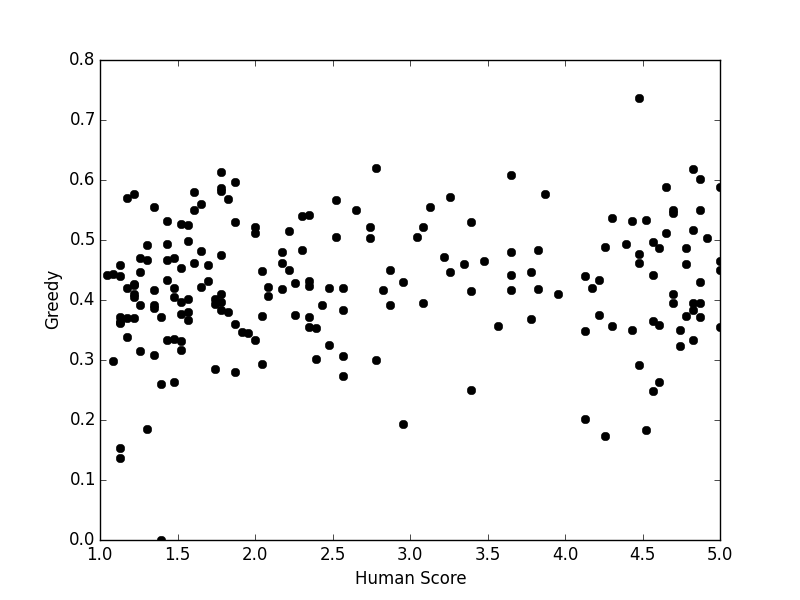}
            \caption{Greedy Matching}
\end{subfigure}\\\begin{subfigure}[b]{1.0\textwidth}
            \centering
            \includegraphics[width=0.45\linewidth]{avg_twitter.png}
            \includegraphics[width=0.45\linewidth]{avg_ubuntu.png}
            \caption{Vector Averaging}
\end{subfigure}\\

\caption{\label{fig:scatteremb} Scatter plots showing the correlation between metrics and human judgement on the Twitter corpus (left) and Ubuntu Dialogue Corpus (right). The plots represent vector extrema (a), greedy matching (b), and vector averaging (c).}
\end{figure*}

\begin{figure*}
\centering

\begin{subfigure}[b]{1.0\textwidth}
            \centering
            \includegraphics[width=0.42\linewidth]{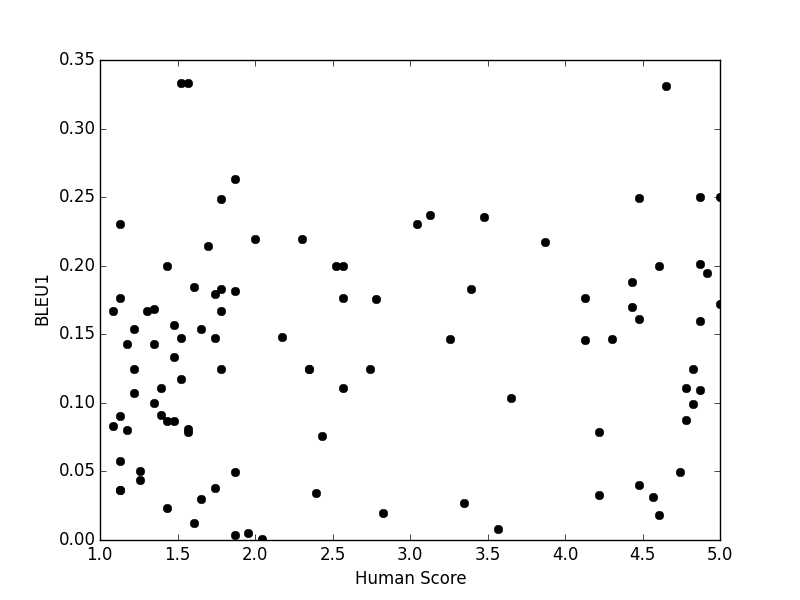}
            \includegraphics[width=0.42\linewidth]{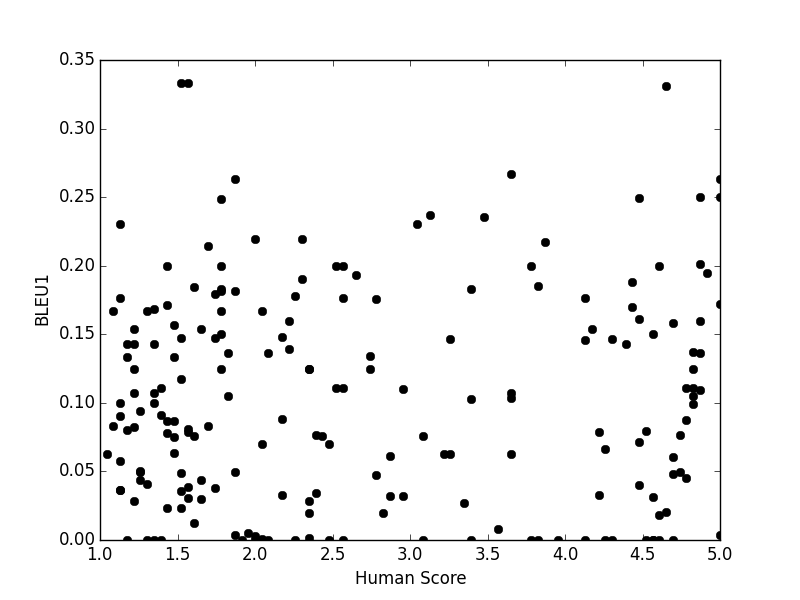}
            \caption{BLEU-1}
\end{subfigure}\\
\begin{subfigure}[b]{1.0\textwidth}
            \centering
            \includegraphics[width=0.42\linewidth]{bleu2_twitter.png}
            \includegraphics[width=0.42\linewidth]{bleu2_ubuntu.png}
            \caption{BLEU-2}
\end{subfigure}\\\begin{subfigure}[b]{1.0\textwidth}
            \centering
            \includegraphics[width=0.42\linewidth]{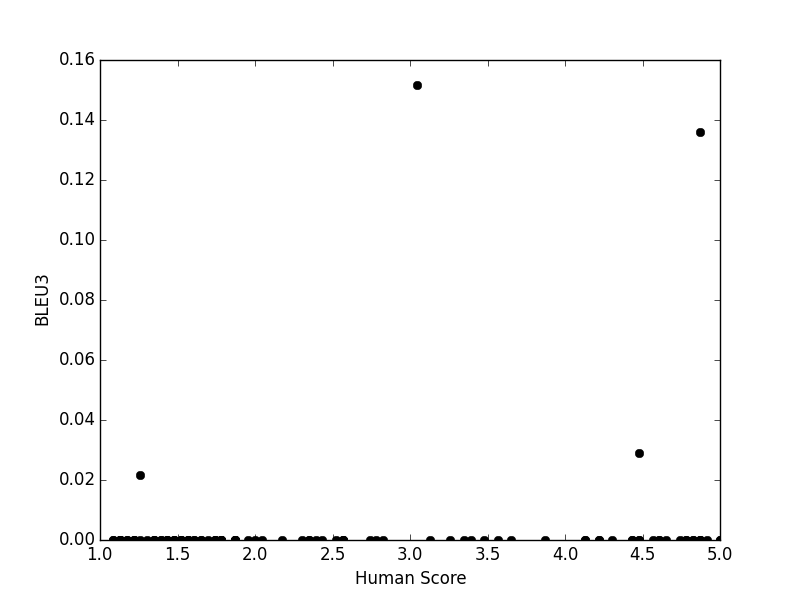}
            \includegraphics[width=0.42\linewidth]{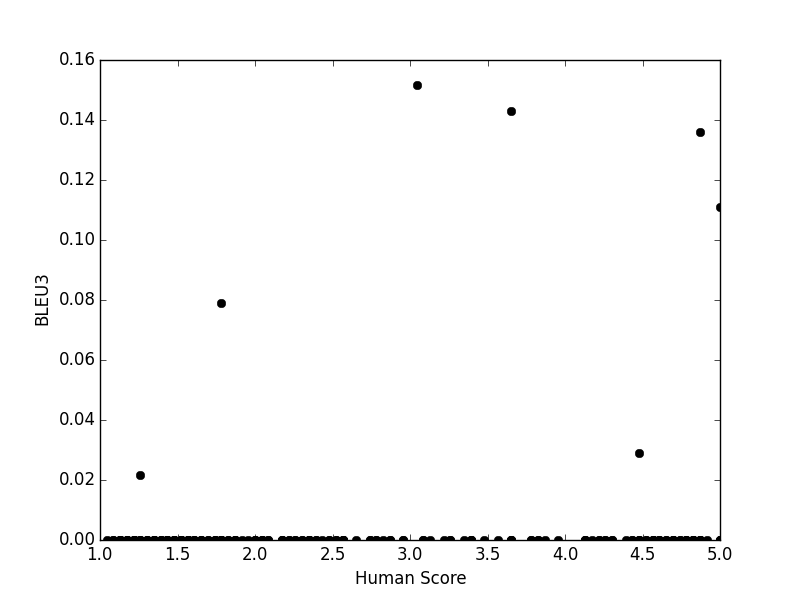}
            \caption{BLEU-3}
\end{subfigure}\\\begin{subfigure}[b]{1.0\textwidth}
            \centering
            \includegraphics[width=0.42\linewidth]{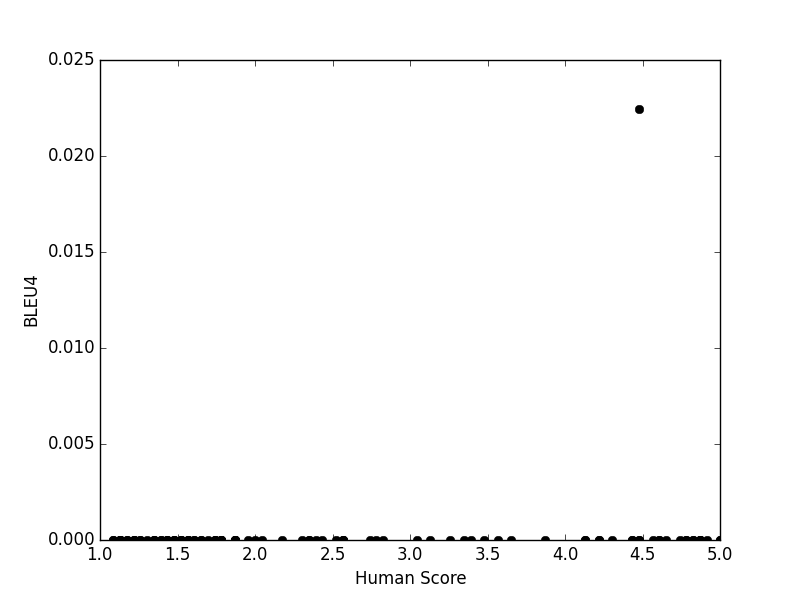}
            \includegraphics[width=0.42\linewidth]{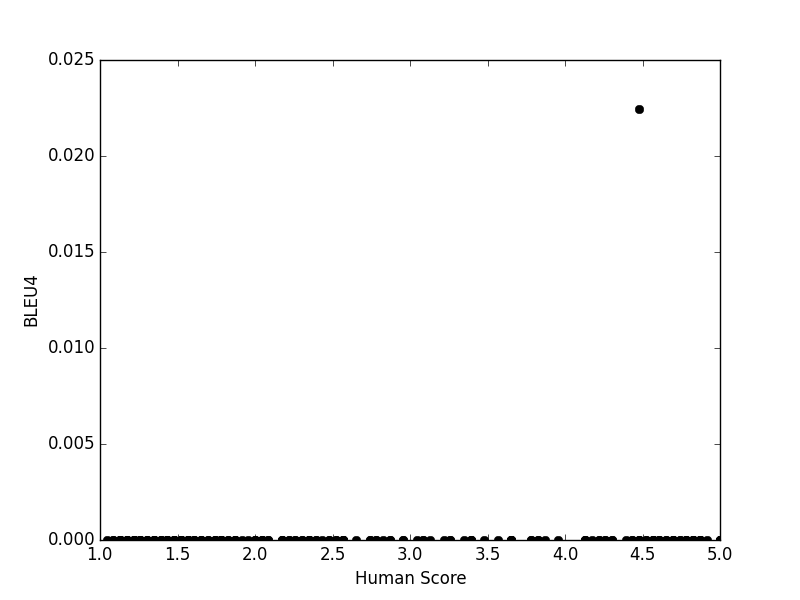}
            \caption{BLEU-4}
\end{subfigure}\\

\caption{\label{fig:scatterbleu} Scatter plots showing the correlation between metrics and human judgement on the Twitter corpus (left) and Ubuntu Dialogue Corpus (right). The plots represent BLEU-1 (a), BLEU-2 (b), BLEU-3 (c), and BLEU-4 (d).}
\end{figure*}

\end{document}